\pdfoutput=1
\documentclass[11pt,a4paper,twocolumn]{article}

\usepackage[utf8]{inputenc}
\usepackage[english]{babel}
\usepackage{csquotes}
\usepackage{amsmath}
\usepackage{textcomp}
\usepackage{preprint}

\usepackage[style=apa, backend=biber, uniquename=false]{biblatex}
\hypersetup{
  pdftitle  = {Concept Representations in Vector Space: Quantifying Sharedness and Idiosyncrasy of Meaning},
  pdfauthor = {Simon Josef Durstewitz, Denis Buehler}
}

\title{Concept Representations in Vector Space:\\
Quantifying Sharedness and Idiosyncrasy of Meaning}

\authors{%
  {\large\textbf{Simon Josef Durstewitz}\affilmark{1}\qquad\textbf{Denis Buehler}\affilmark{2}}\\[0.6em]
  {\small\affilmark{1}Copenhagen Center for Social Data Science (SODAS), University of Copenhagen, Denmark}\\
  {\small\affilmark{2}Institut Jean Nicod, École normale supérieure -- PSL, Paris, France}\\[0.3em]
  {\small\href{mailto:cjm902@sodas.ku.dk}{\texttt{cjm902@sodas.ku.dk}}\qquad\href{mailto:denis.buehler@ens.psl.eu}{\texttt{denis.buehler@ens.psl.eu}}}%
}

\newcommand{\osfdata}{\href{https://osf.io/dc6an/}{osf.io/dc6an}}
\newcommand{\osfprereg}{\href{https://osf.io/czxqe/}{osf.io/czxqe}}

\preprintnote{Preprint under review at \textit{Open Mind: Discoveries in Cognitive Science}. 
Part of this work was conducted during a research stay at the Institut Jean Nicod, \'Ecole normale sup\'erieure -- PSL.
Correspondence: Simon Josef Durstewitz (\href{mailto:simon.durstewitz@gmail.com}{simon.durstewitz@gmail.com}). Data \& code: \osfdata. Pre-registration: \osfprereg.}

\begin{document}

\maketitle

\begin{abstract}
Concepts are the basic units through which we navigate the world and communicate meaning. But do you and I really mean the same thing when we think or speak about a concept? The present study develops a measurement framework for modeling and analyzing interindividual variation in concept representations through sentence embeddings, dense vector representations mapping entire utterances into a shared geometric space. The framework is developed in a theoretical argument and validated, demonstrated, and subjected to robustness checks in a pre-registered study. Short personal definitions, elicited from $N = 125$ participants, are encoded as sentence embeddings and submitted pairwise to a crossed random effects model yielding two concept-level quantities: sharedness ($S_b$), the expected dissimilarity between two randomly drawn individuals, and idiosyncrasy ($I_b$), the person-specific share of total dyadic variance. Two validity tests support the approach: the embedding space fully recovers concept and domain organization from participant definitions even with the target word excluded, and a parameter-free classifier achieves 0.630 accuracy in a within-participant extensional prediction task against a chance baseline of 0.500 and a consistency ceiling of 0.893. The framework finds theoretically interpretable variation in $S_b$ and establishes $I_b$ as a reliable source of dyadic variance across concepts. Results replicate across three encoders varying in architecture and capacity, providing a reproducible and scalable instrument for the empirical study of meaning variation.
\end{abstract}

\section{Introduction}
Concepts are the ``building blocks of thought’’ \parencite[p.~1]{margolis_concepts_2023}. They underpin cognitive processes such as categorization, inference, memory, decision-making and action planning, and enable people to coordinate meaning in communication \parencite{hampton_investigating_2020}. But do different individuals mean the same thing when they invoke the same concept?

Empirical evidence has shown that individuals often grasp the same concept differently, maintaining distinctive personal representations \parencite{hampton_investigating_2020}. While considered key to understanding how individuals navigate the world \parencite{von_glasersfeld_radical_1995}, concept representations, as latent mental structures, are difficult to measure \parencite{petrenco_centroid_2026}, and the question of whether individuals converge in the private meanings they hold for a concept remains ``largely underexplored'' \parencite[p.~5]{visconti_di_oleggio_castello_individual_2025}.

The present study draws on advances in natural language processing (NLP) to develop a measurement framework that models individual concept representations in vector space, opening these up for quantitative comparison. Short, standardized personal definitions of concepts, elicited from participants, are encoded as sentence embeddings \parencite{reimers_sentence-bert_2019} and compared geometrically. This modeling approach allows the question of meaning variation to be addressed statistically from two perspectives, (1) how shared representations of a concept are across individuals and (2) how much of the variance in divergence is accounted for by individual distinctiveness. While prior work has used textual traces as an empirical route to concept representations \parencite{bellezza_reliability_1984, bellezza_reliability_1984-1, petrenco_centroid_2026, aeschbach_mapping_2025, johns_determining_2024}, the present design responds to three shortcomings of existing approaches. First, reliance on naturalistic text confounds differences in concept representations with context of language use. Second, token-level representations discard compositional and relational structure central to externalizing concept representations in language. Third, existing methods treat the group or dyadic mean as the primary unit of analysis, precluding individual-level inference.

The framework is developed in a theoretical argument and demonstrated in a pre-registered study, including validity and robustness checks. The paper begins by reviewing the existing literature, turning first to the empirical record against which this framework is situated, and then to existing measurement approaches and the gaps that motivate the present study.

\section{Theoretical Background}
A long-standing question concerns whether concepts are universal or, in some sense, personal \parencite{murphy_how_2020}. Philosophical tradition has often treated concepts as public and objective, shared across individuals by definition \parencite{margolis_concepts_2003} but empirical research shows that individuals differ markedly in both how they understand a concept and which entities they take it to include \parencite{hampton_investigating_2020}. 

The debate has proven difficult to resolve because positions diverge along theoretical fault lines, some construing concepts as abstract and mind-independent \parencite{frege_gedanke_1918}, others as externally determined \parencite{putnam_what_1975}, and still others as internal mental information structures \parencite{margolis_concepts_2023}. The difficulty partly dissolves once two levels of analysis are distinguished. Regardless of whether concepts exist as mind-independent entities of some sort, individuals associate a private understanding or ``conception'' \parencite[][p.~317]{burge_concepts_1993} with them in some internal format that can be partial or mistaken while still shaping an individual's inferences and behavior \parencite{burge_individualism_1979}.

The present study terms such private understandings \textit{concept representations}, denoting the mental information structures through which individuals encode and organize their understanding of concepts \parencite[][]{gardenfors_conceptual_2004, jackendoff_what_1992}. This allows the question to be reframed from binary--are concepts universal or not?--to continuous: for a given concept, to what extent do individuals’ representations of it converge? 

The question has far-reaching implications, as concept representations underpin processes across multiple levels. At the level of the individual, this could explain differences in how people categorize, draw inferences, and learn \parencite[e.g.,][]{djalal_relationship_2018}. In social interactions, individuals bring their own private meanings to the exchange. Divergence can lead to misunderstandings in communication \parencite{von_glasersfeld_radical_1995}, and may offer an underappreciated explanation for failures of mutual understanding \parencite{marti_latent_2023}. Divergence may carry consequences on an even greater scale when it concerns institutional facts such as money or citizenship, which are ontologically dependent on sufficient agreement \parencite{searle_construction_1995}, and could otherwise become fragile. Another area currently coming into focus with the rise of AI systems is the alignment between human and model representations of concepts such as fairness, harm, or truth, seen as critical for safe deployment \parencite{gabriel_artificial_2020}.

\subsection{Variation Between Individuals}
 
Concept representations are typically understood as comprising an intension and an extension \parencite{hampton_investigating_2020}. The intension is the set of properties characteristic of the concept determining category membership. The extension is the set of entities in the world to which the concept applies. Together, they form an interpretive structure through which individuals navigate, communicate, and reason about the world, binding the stream of experience to stored knowledge \parencite{gardenfors_conceptual_2004} and enabling generalization to novel instances \parencite{barsalou_human_2012}.
 
Interindividual variation in concept representations is well documented across experimental tasks. Between-participant agreement is low in judgments of what falls under a concept \parencite{mccloskey_natural_1978}, what defines it \parencite{bellezza_reliability_1984-1}, even among domain experts \parencite{levitis_behavioural_2009}, as well as in exemplar generation \parencite{bellezza_reliability_1984} and typicality judgments \parencite{barsalou_instability_1987}. Evidence from large-scale text corpora points the same way. \textcite{johns_determining_2024} derives individualized distributional models of semantic memory from Reddit data and finds that alignment between individuals is low.

While disagreement is consistently found, individuals are largely unaware of this variation and exhibit a strong bias toward assuming that others share their understanding \parencite{marti_latent_2023}.

\subsection{Variation Across Concepts}

The degree to which representations are shared plausibly varies across concepts, reflecting differences in their properties. Concepts differ in how their content is constrained by the structure of the world \parencite{atran_basic_1989}. Natural kinds are often accorded a privileged status because the categories they define are assumed to reflect the structure of the natural world rather than human interests or conventions \parencite{bird_natural_2022, hacking_tradition_1991, cohnitz_variation_2020}, plausibly stabilizing representations through shared physical referents, sensorimotor grounding \parencite{wang_fmri_2022}, and psychological essentialism \parencite{gelman_psychological_2004}. Social and abstract concepts, by contrast, derive more of their content from collective intentionality and convention \parencite{searle_construction_1995} and from language encounters rather than perceptual experience \parencite{lenci_emotions_2018, solovyev_concretenessabstractness_2021}, plausibly affording greater representational divergence. Empirical work finds initial support for this argument, with neural evidence showing that social concepts account for a disproportionate share of differences in brain representations \parencite{visconti_di_oleggio_castello_individual_2025}. Other work documents 
a concreteness gradient in which variability decreases with perceptual grounding and increases with abstractness \parencite{wang_idiosyncratic_2021, marti_latent_2023}.

A related dimension concerns the position of concepts in a taxonomic hierarchy. Following \textcite{rosch_principles_1978}, concepts can be situated at three levels: superordinate concepts are broad and heterogeneous with few shared attributes (e.g., fruit); basic-level concepts represent an intermediate level (e.g., apple); subordinate concepts are specific, with members sharing most attributes with contrasting subordinates (e.g., Granny Smith). Representations have been shown to converge most strongly at the basic level, exhibiting greater definitional agreement relative to superordinate concepts \parencite{bellezza_reliability_1984-1} and higher feature convergence relative to both superordinate and subordinate levels \parencite{zhuang_characterization_2022}.

\subsection{Personal Identity as a Source}

Individuals encode their understanding of concepts through personal experience \parencite{margolis_learning_2011, clausner_domains_1999}, both through direct sensorimotor encounters with concept referents and, for more abstract concepts, through linguistically mediated acquisition \parencite{lenci_emotions_2018, solovyev_concretenessabstractness_2021}. This suggests that representations reflect a biographical history of encounters that is unique to each person \parencite{von_glasersfeld_radical_1995}, thereby pointing to personal identity as a source of the variation documented above.

Evidence is consistent with this account. A range of studies finds within-participant agreement substantially exceeding between-participant agreement \parencite{bellezza_reliability_1984, bellezza_reliability_1984-1, barsalou_instability_1987, johns_determining_2024, djalal_relationship_2018}, indicating that individuals are more similar to themselves than to others. \textcite{hampton_investigating_2020} takes this as ``prima facie evidence'' (p.~70) for stable personal representations, termed \textit{idio-prototypes}, that differ systematically from one person to another, while acknowledging methodological and theoretical uncertainties. Neurological evidence converges on this account, finding qualitatively distinct brain representations for the same concepts across individuals \parencite{wang_idiosyncratic_2021}, with person-specific signatures explaining over 30\% of variance within a participant but less than 2\% in others \parencite{visconti_di_oleggio_castello_individual_2025}.

While theoretical and empirical accounts converge on the existence of stable personal representations, a continuous estimate of the extent to which individual distinctiveness accounts for divergence is lacking, partly because existing approaches rely on mean pairwise statistics that obscure person-level structure \parencite{djalal_relationship_2018}.

\subsection{Language and the Study of Representational Variation}
The methods available determine ``the field accessible for scientific investigation'' \parencite[p.~78]{kuhn_structure_1996}, and the study of concept representations is no exception. Language has long served as the primary instrument through which representations are accessed, and textual data have been analyzed with increasing methodological sophistication.

Early studies decomposed participant definitions into discrete propositions \parencite{bellezza_reliability_1984-1} or relied on semantic feature listing protocols \parencite{bellezza_reliability_1984}, computing comment–element correlations based on binary match–no-match logic. Although informative at the time, the approach is labor intensive, and the binary matching logic fails to capture graded semantic variation such as synonymy or partial semantic overlap.

Distributional semantic models represent word meanings as high-dimensional vectors derived from co-occurrence statistics, enabling graded estimates of semantic similarity at a scale inaccessible to manual methods. Building on this approach, \textcite{johns_determining_2024} constructs person-specific count-based distributional matrices from personal Reddit corpora across thousands of words to compute between-person semantic similarity. Three limitations constrain this approach, however. First, reliance on naturalistic text confounds representational with contextual variation, as users self-select into topical communities and adapt their language to audience, register, and genre, such that differences in word use need not reflect differences in underlying representations. Second, count-based models require large text corpora to produce stable vector representations. Third, aggregating co-occurrence statistics does not capture the compositional structure of linguistic expressions and leaves lexical ambiguity unresolved \parencite{turney_frequency_2010}.

Recent approaches use large-scale free-association data, eliciting several associates per cue word, to analyze representational differences while avoiding many confounds of naturalistic text. \textcite{aeschbach_mapping_2025} transform associates into embeddings based on response-distribution statistics or language-model feature extraction and compare group-specific dissimilarity matrices, making the group rather than the individual the unit of analysis. \textcite{petrenco_centroid_2026} extend this approach to the individual by averaging word embeddings into a participant centroid, but find individual-level estimates substantially less accurate than group-level estimates and leave the analysis of interindividual variation to future work. Both approaches operate on isolated associates, however, also discarding compositional and relational structure and leaving lexical ambiguity unresolved. The use of static word embeddings further collapses variation across speakers and contexts into a single decontextualized vector per word.

A historical overview shows that, while advances in text representation have enabled graded measures of semantic similarity and analyses at much larger scale, central challenges remain in (1) avoiding the confounds of naturalistic text through standardized elicitation, (2) stably modeling the individual as the unit of analysis, and (3) preserving compositional and relational structure. Advances in transformer-based NLP make it possible to address these challenges within a unified framework.

\section{A New Measurement Framework}
\label{sec:framework}
\begin{figure*}[t]
\centering
\includegraphics[width=0.72\textwidth]{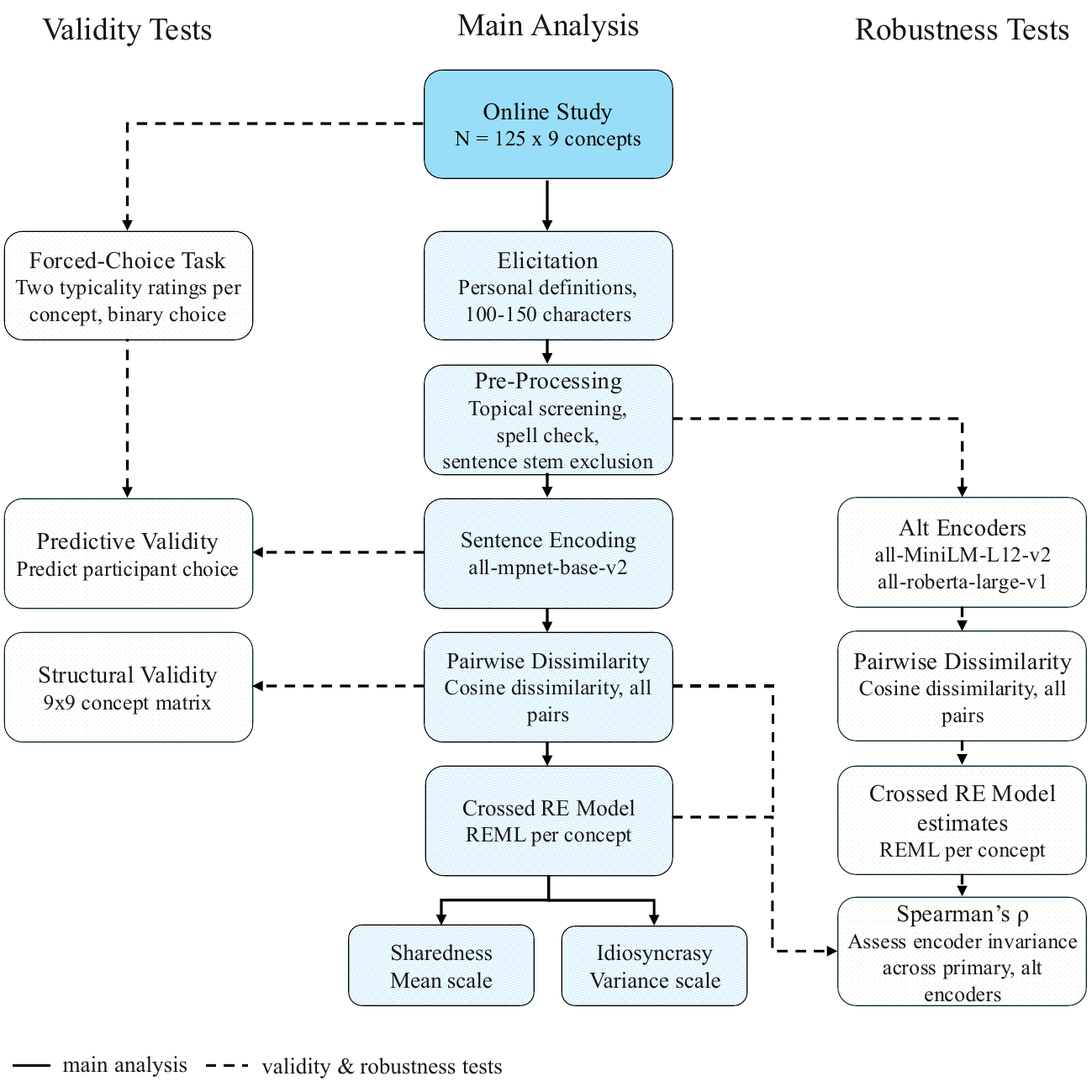}
\caption{The center column shows the main analysis, the left column reports validity assessments, and the right column robustness checks using alternative (Alt) encoders.}
\label{fig:approach}
\end{figure*}

The present study proposes a measurement framework that advances the study of conceptual variation in two ways: first, by encoding participant responses as transformer-based sentence embeddings; and second, by introducing a statistical framework that measures the sharedness of representations on a common scale across concepts and formalizes individual distinctiveness as a continuous and comparable source of variation (Figure~\ref{fig:approach}).

Sentence embeddings encode entire utterances as dense high-dimensional vectors, interpreting each token relative to all others through self-attention and thereby generating contextualized representations that preserve compositional and relational structure~\parencite{vaswani_attention_2017}. Fine-tuned on semantic similarity tasks, they are further calibrated to human similarity judgments, placing semantically similar utterances in proximate regions of the embedding space \parencite{reimers_sentence-bert_2019}. This form of representation addresses the previously identified challenges in several respects. 

First, sentence encoders capture graded semantic similarity between texts and can produce stable representations even from short textual inputs \parencite{xu_revealing_2025}, making the approach compatible with standardized elicitation paradigms of earlier experimental work while permitting stable modeling at the individual level. Second, by operating on full sentences rather than isolated tokens, they are better suited to capture real-world modulation of meaning, while training on semantic similarity further reduces sensitivity to stylistic variation \parencite{reimers_sentence-bert_2019}. Finally, the approach situates all participant representations within a shared vector space, enabling direct comparison, visualization, and quantitative analysis across concepts and individuals.

Obtained vector representations are then modeled statistically as dyads in a crossed random effects (CRE) model. Prior approaches summarize pairwise distances as a mean, treating all dyads as interchangeable and discarding information about who contributes to which pairing \parencite[e.g.,][]{bellezza_reliability_1984, bellezza_reliability_1984-1, johns_determining_2024}. The present study retains this information by taking mathematical inspiration from the Social Relations Model \parencite{kenny_dyadic_2011, kenny_social_1984}, yielding two estimands. 

Sharedness ($S_b$) is the expected dissimilarity between two randomly drawn individuals, estimated as a fixed intercept, and captures the degree of mean representational convergence. Idiosyncrasy ($I_b$) is the person-specific share of total dyadic variance, estimated from the consistency of each individual's distances across all dyads in which they appear. While $S_b$ quantifies how much representations diverge across individuals, $I_b$ quantifies whether that divergence is person-structured---i.e., whether individual distinctiveness is a consistent predictor of dyadic distance---lending empirical grounding to \textcite{hampton_investigating_2020}'s notion of idio-prototypes.

\subsection{Rationale}
The framework proposes to elicit concept representations in linguistic format, encode textual traces in embedding space, and compare vector representations geometrically. The following makes explicit the theoretical commitments underlying each level, formulated as three premises (P1--P3).

\begin{figure*}[t]
\centering
\includegraphics[width=0.85\textwidth]{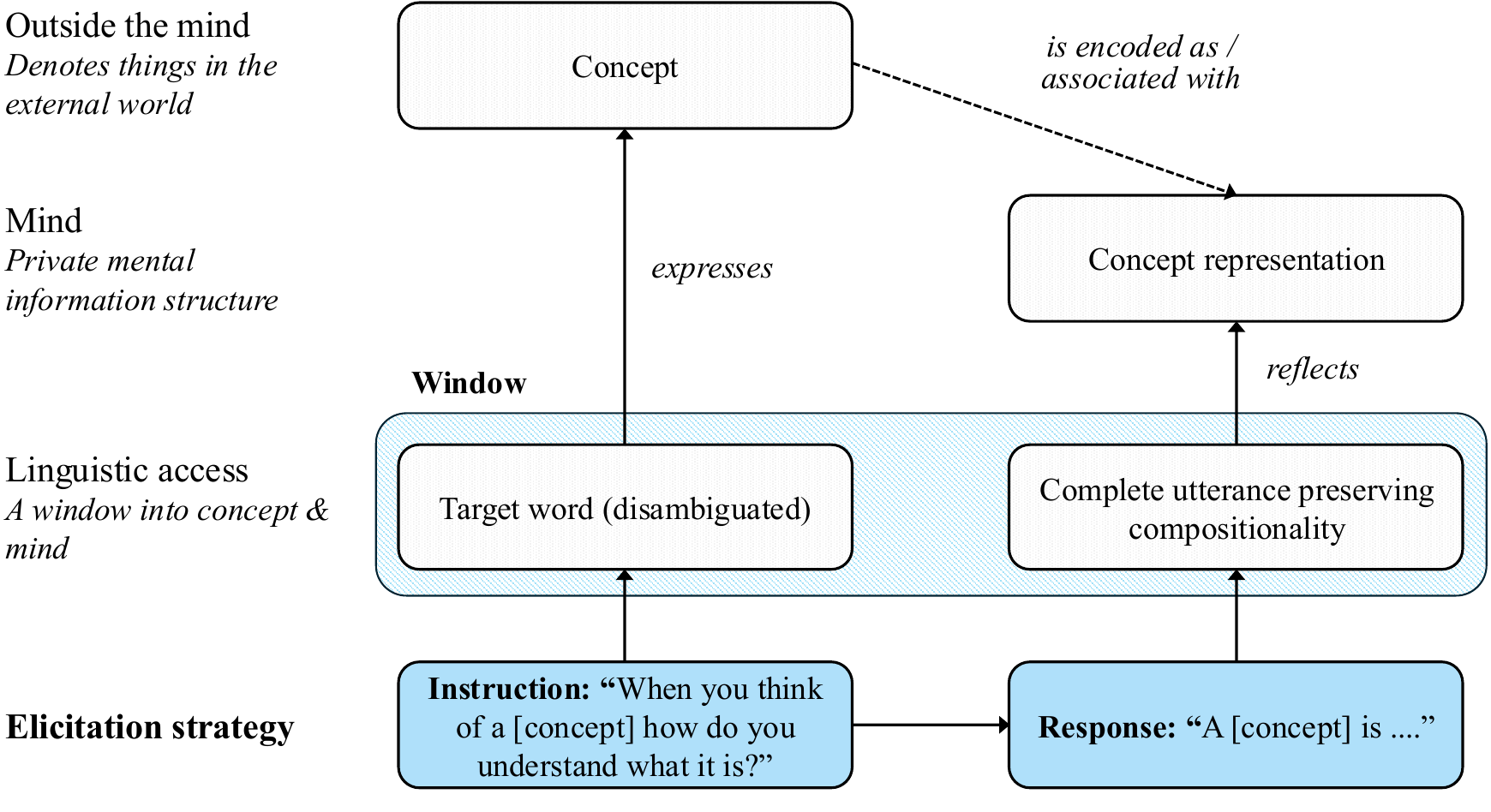}
\caption{The illustration comprises four levels, from top to bottom: outside the mind, mind, linguistic access, and elicitation strategy. A target word is taken to express a concept and representations are then elicited in linguistic format to empirically open these up for comparison.}
\label{fig:premise}
\end{figure*}

\subsubsection{(P1) Language provides a window into concept and mind.}  Central to the framework is the role of language, both (1) in the instruction and (2) in the response format (Figure~\ref{fig:premise}). In the instruction, a target word expresses a concept \parencite{frege_uber_1892}, for which a concept representation is then elicited. While single words are often polysemous, mapping onto multiple concepts, linguistic context can resolve such ambiguity, in which case the word is assumed to be understood as expressing the same concept across participants.

In the response format, this study draws on a long tradition that understands language as a “mirror of mind” \parencite[von Humboldt, as cited in][p.~67]{chomsky_language_2006} and, more specifically, as “a window on conceptual structures” \parencite[][p.~77]{von_glasersfeld_radical_1995}. Although language and mental representations are generally taken as distinct levels of analysis \parencite{malt_representing_2024}, language is argued to systematically relate to and provide privileged access to mental representations. Words, on this view, function as signs that point to underlying representations \parencite{ogden_meaning_1923, von_glasersfeld_radical_1995}. Beyond isolated words, the productivity and compositionality of language are argued to mirror the structure of thought \parencite{fodor_language_1975}, and compositional arrangement in particular is argued to make complex aspects of a concept's representation explicit, such as its internal relational structure \parencite{jackendoff_languages_1992}.

Accordingly, participants are assumed to externalize concept representations in complete utterances, where compositionality can manifest, bottom-up and without constraint by the researcher's prior assumptions \parencite{petrenco_centroid_2026, aeschbach_mapping_2025}, such that the resulting textual traces reflect systematic differences in underlying representations rather than mere surface-level noise.

\subsubsection{(P2) Embedding spaces preserve semantic structure.} Embedding spaces are high-dimensional vector spaces that represent linguistic input. Their dimensions capture statistical regularities in language \parencite{bengio_neural_2003}, operationalizing the hypothesis that semantic structure can be inferred from word--context co-occurrence patterns \parencite{harris_distributional_1954, firth_synopsis_1957}. While individual dimensions are not interpretable, they jointly encode semantic relations in geometric structure, such that semantically similar input is positioned closer together \parencite{mikolov_efficient_2013, mikolov_linguistic_2013}.

Embedding spaces are often interpreted as reflecting psychologically plausible learning principles and as proxies of human cognitive organization \parencite{gunther_vector-space_2019}, a view supported by empirical evidence. Embeddings predict performance on word association, similarity, and analogy tasks \parencite{pereira_comparative_2016}, account for higher-level behavioral outcomes such as probability judgments and consumer choice \parencite{bhatia_distributed_2019}, and support the geometric recovery of detailed human knowledge about perceptual and object features \parencite{carter_leveraging_2025, grand_semantic_2022}.

Different architectures underlying embedding spaces can be distinguished. Static models such as LSA or GloVe assign each word type a single, fixed vector \parencite{landauer_introduction_1998, pennington_glove_2014, mikolov_efficient_2013}. Transformer-based models \parencite{vaswani_attention_2017} instead compute a representation for each token using self-attention, which weights every token against every other token in the input, so that the same word receives different representations in different contexts \parencite{wang_language_2019}. Fine-tuning Transformers on semantic similarity data produces sentence embeddings, single vectors for complete utterances, trained so that semantically similar sentences are placed close together in vector space \parencite{reimers_sentence-bert_2019}.

The framework accordingly takes sentence embeddings as a principled measurement instrument, positioning textual traces as points in a shared vector space for geometric analysis.

\subsubsection{(P3) Relations between concept representations can be examined in geometric terms.}
The idea that similarity or distance in a property space is an organizing principle of conceptual structure has a long history, with early roots found in Hume's principle of association by resemblance \parencite{hume_treatise_1739}. Empirical work using multidimensional scaling gave this intuition a formal basis, demonstrating that mental representations can be recovered as points in a metric space in which geometric distance aligns with human similarity judgments \parencite{shepard_multidimensional_1980}. This perspective is further developed in \textcite{gardenfors_conceptual_2004}'s theory of conceptual spaces, which represents concepts as regions in multidimensional quality spaces, where similarity is captured by geometric proximity.

More recently, \textcite{piantadosi_why_2024} argue that concepts can be understood as points in high-dimensional vector spaces, showing that such an account can accommodate seemingly competing theories of concepts, including classical, prototype, exemplar, relational, and theory-theory accounts. 

Without committing to stronger ontological claims, the framework takes this as motivation for treating similarity between concept representations as a spatial relation. Represented as points in an embedding space, textual traces are compared using cosine similarity, as Euclidean distance becomes unreliable in high-dimensional spaces \parencite{aggarwal_surprising_2001}.

With the theoretical rationale in place, the following section turns to an empirical demonstration of the framework.

\section{Method}
\label{sec:method}
An online survey study was pre-registered prior to data collection, specifying the target sample size, analysis plan, and primary measurement decisions (see \osfprereg). Its aims were to validate, demonstrate, and assess the robustness of the framework on a purposive sample of nine concepts.

\subsection{Concept Selection}

The concepts included in the study were selected along two theoretically motivated dimensions. The first is conceptual domain. Three domains were selected: natural types, artefacts, and social concepts. These differ in the basis of their defining features: natural kinds by perceptual and biological properties, artefacts by agentive functions, and social concepts by collectively sustained functions. Prior work documents representational divergence across domains \parencite{marti_latent_2023, wang_idiosyncratic_2021}, suggesting domain-level variation in $S_b$ and $I_b$. The second is taxonomic level \parencite{rosch_principles_1978}. Empirical work has shown that representations converge most strongly at the basic level \parencite{bellezza_reliability_1984-1, zhuang_characterization_2022}, motivating the examination of $S_b$ and $I_b$ across taxonomic levels.

\begin{figure*}[t]
\centering
\includegraphics[width=0.72\textwidth]{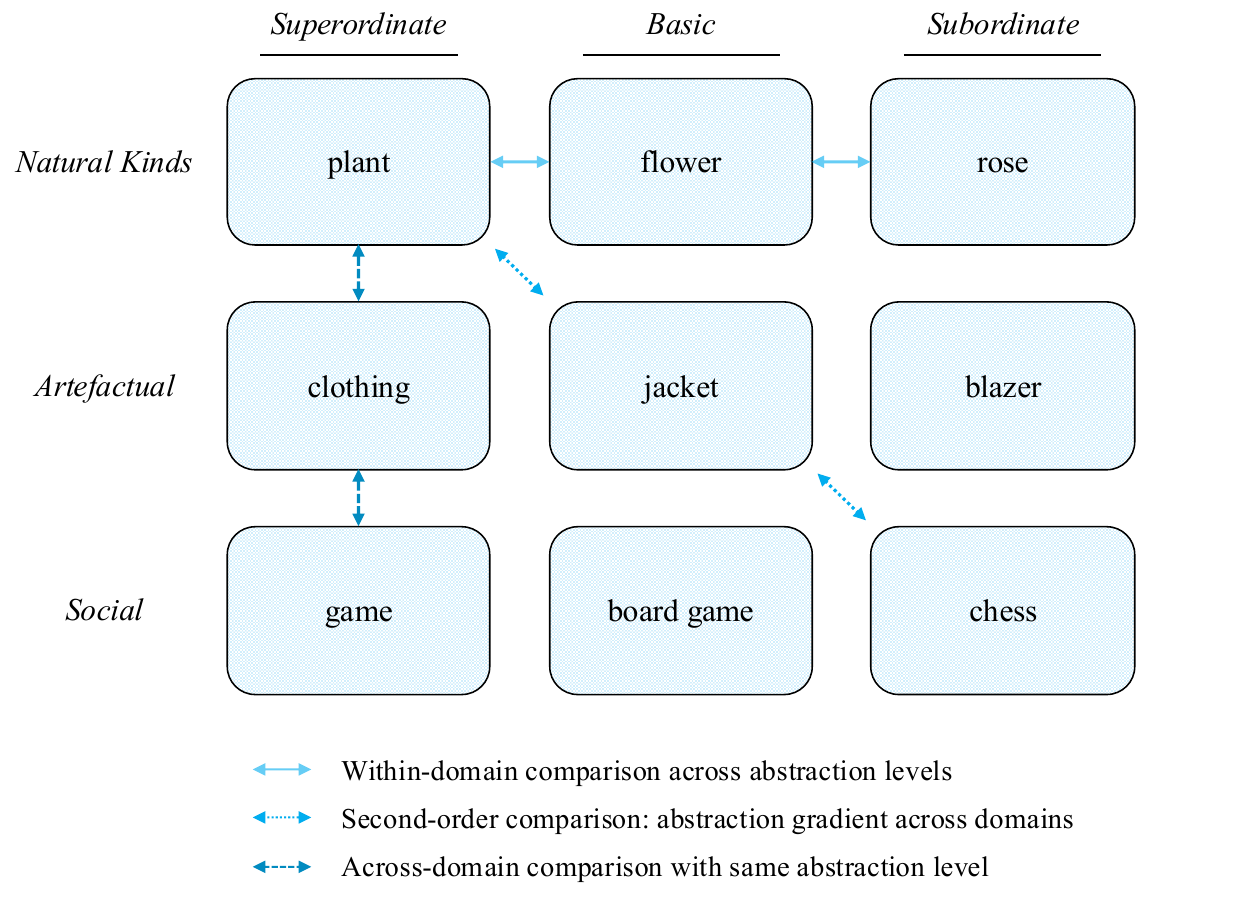}
\caption{Arranging the concept set along two dimensions affords a comparatively rich set of analytical comparisons from a small number of concepts. Within-domain and second-order comparisons are also less susceptible to possible non-uniform semantic resolution (see also Section~\ref{sec:limitations}). Note, however, that no comparison supports causal attribution, as each cell contains only one concept, and concepts may differ on more than the controlled dimensions. The purpose of the demonstration is to surface and explore variation.}
\label{fig:design}
\end{figure*}

Crossing the three taxonomic levels with the three domains yields a 3$\times$3 design, with one concept per cell (Figure~\ref{fig:design}). Concepts were selected to be everyday, widely familiar, and largely non-polysemous. Presenting them with an indefinite article further disambiguates them by fixing their part of speech as a noun.

\subsection{Participants}

A minimum target of $N = 116$ was determined heuristically, a priori, by two criteria: stable estimation of variance components in CRE models based on simulation evidence \parencite{chung_impact_2018} and precision of the forced-choice accuracy estimate, with the latter being the more conservative binding constraint. Full derivation is provided in the pre-registration. 

Participants were recruited through convenience sampling via university mailing lists, physical postings, and online communities, supplemented by Prolific to meet the minimum target within the data collection window. Eligibility required age 18 or older, informed consent, and English proficiency at C1 level or above, assessed via screening items at survey entry. Participation was incentivized through a lottery offering seven prizes of \texteuro{10} each and monetary compensation for Prolific participants. Data collection ran from March 2 to April 24, 2026. The realized sample comprised $N = 125$ valid responses ($M_{\text{age}} = 25.9$, $SD = 9.1$; 58.4\% women; 80.8\% non-native English speakers) drawn from 18 countries. Full sociodemographic distributions are reported in Table~\ref{tab:sociodemographic}.

\subsection{Procedure}
The study was administered as an online survey using SoSci Survey \parencite{leiner_sosci_2026} and comprised three blocks: (1) sociodemographic information, (2) a definition task, and (3) a forced-choice task. The order of blocks 2 and 3 was randomized between participants; item order within each block was randomized independently. Participants provided informed consent prior to participation and were informed of their right to withdraw at any time without consequence. 

\subsubsection{Sociodemographic Information.} Participants responded to the items summarized in Table~\ref{tab:sociodemographic}. All items were optional.

\subsubsection{Definition Task.} Participants provided personal definitions of nine concepts in response to a standardized prompt: \textit{``When you think of [concept], how do you understand what it is? Please describe it in your own words.''}, with each field prefilled by a concept-specific sentence stem (e.g., \textit{A plant is\ldots}) that participants completed. The stem imposed a uniform syntactic frame, standardizing inputs and eliciting declarative descriptions, and was later removed prior to embedding. Responses were constrained to 100--150 characters to ensure comparability across participants, guarantee sufficient semantic content, and prevent padding that shifts the embedding away from the participant's intended meaning. A color-coded character counter facilitated compliance and pasting text was disabled to ensure self-generated responses.

\subsubsection{Forced-Choice Typicality Task.} A forced-choice typicality task covered the same nine concepts as the definition task, consisting of 18 items, two per concept, presented as a separate block (Table~\ref{tab:closedend}). Response option order was randomized between participants. Three items were randomly drawn and repeated to assess within-participant response consistency, and one attention check item was included. On each item, participants were shown two researcher-constructed descriptions that instantiated the concept at a more specific level, and chose the one that better matched their own understanding. For example, for \textit{plant}, participants chose between ``a tall oak tree rising from the ground in a wild forest'' and ``a small ornamental orchid kept indoors in a pot.'' 

Response options appeared after a three-second delay during which participants were prompted to think of the concept, ensuring their own representation was activated prior to exposure to the options. Whereas participants defined concepts at a general, intensional level in the definition task, the descriptions here instantiate lower-order characterizations of each concept, such that selecting the more typical description requires extensional inference.

\subsection{Data Processing and Encoding}

At the participant level, responses were excluded for attention check failures. To address the growing threat of LLM-generated responses in survey research \parencite{westwood_potential_2025}, a honeypot field was included to detect automated submissions, with flagged responses excluded. No submissions triggered the field. 

At the definition level, three pre-processing steps were applied. First, definitions were screened for topical coherence, excluding those lacking interpretable semantic content or clearly unrelated to the target concept while retaining borderline cases. Thirteen definitions were excluded and are documented in the supplementary materials. Second, spelling was corrected to prevent degraded tokenization from conflating orthographic and representational variation \parencite{nishimwe_making_2024}. Errors were flagged computationally and corrected manually, limited to unambiguous misspellings while retaining informal but interpretable forms. Third, the sentence stem shared across responses to a given concept was removed so that definitions captured only participant-generated content.

Pre-processed definitions were encoded as sentence embeddings using the SBERT framework \parencite{reimers_sentence-bert_2019}. Encoder selection was informed by Semantic Textual Similarity (STS) benchmarks \parencite{enevoldsen_mmteb_2025}, and \texttt{all-\allowbreak mpnet-\allowbreak base-\allowbreak v2} was selected as the primary encoder, a state-of-the-art general-purpose encoder with strong benchmark performance \parencite{song_mpnet_2020}. 

\subsection{Main Analysis}

The unit-normalized embeddings $\mathbf{v}_{ib}$ produced by \texttt{all-mpnet-base-v2} served as the basis for the main analysis, where $i = 1, \ldots, N$ indexes participants and $b = 1, \ldots, 9$ indexes concepts. For each concept $b$, the cosine dissimilarity between every pair of participants $i$ and $j$ was computed as:

\begin{equation}
\delta_{ijb} = \frac{1 - \mathbf{v}_{ib}^\top \mathbf{v}_{jb}}{2} \in [0, 1]
\end{equation}

where the inner product therefore equals the cosine similarity and division by 2 maps the result to $[0, 1]$. This yields $N(N-1)/2$ unordered pairs per concept, a round-robin structure in which every definition is compared against every other. These pairwise dissimilarities are submitted to a CRE model fitted separately for each concept via restricted maximum likelihood:
\begin{equation}
    \delta_{ijb} = \alpha_b + a_{i,b} + a_{j,b} + \varepsilon_{ijb}
\end{equation}
where $\alpha_b$ is the fixed intercept, $a_{i,b}$ and $a_{j,b}$ are participant-specific random effects for the two members of each dyad, and $\varepsilon_{ijb}$ is the residual. The crossed structure is required because each participant appears as both $i$ and $j$ across dyads. The model yields two estimands. The fixed intercept $\alpha_b$ is the expected dissimilarity between two randomly drawn participants, against which \textit{sharedness} is defined as:
\begin{equation}
    S_b = 1 - \alpha_b
\end{equation}
A high $S_b$ indicates that representations of concept $b$ tend to be similar across the population. 

The random effects $a_{i,b}$ and $a_{j,b}$ are assumed i.i.d.\ with variance $\tau_b^2$, quantifying how much stable individual distinctiveness contributes to dyadic variance. The residual $\varepsilon_{ijb}$ has variance $\sigma_b^2$, such that total dyadic variance is $V_b^{\text{tot}} = 2\tau_b^2 + \sigma_b^2$, against which \textit{idiosyncrasy} is defined as:
\begin{equation}
    I_b = \frac{2\tau_b^2}{V_b^{\text{tot}}}
\end{equation}
A high $I_b$ indicates that individual distinctiveness is a consistent predictor of dyadic distance, such that some participants are reliably more distant from all others. Because $S_b$ operates on the mean scale and $I_b$ on the variance scale, the two can be orthogonal. 

\begin{figure*}[t]
\centering
\includegraphics[width=0.72\textwidth]{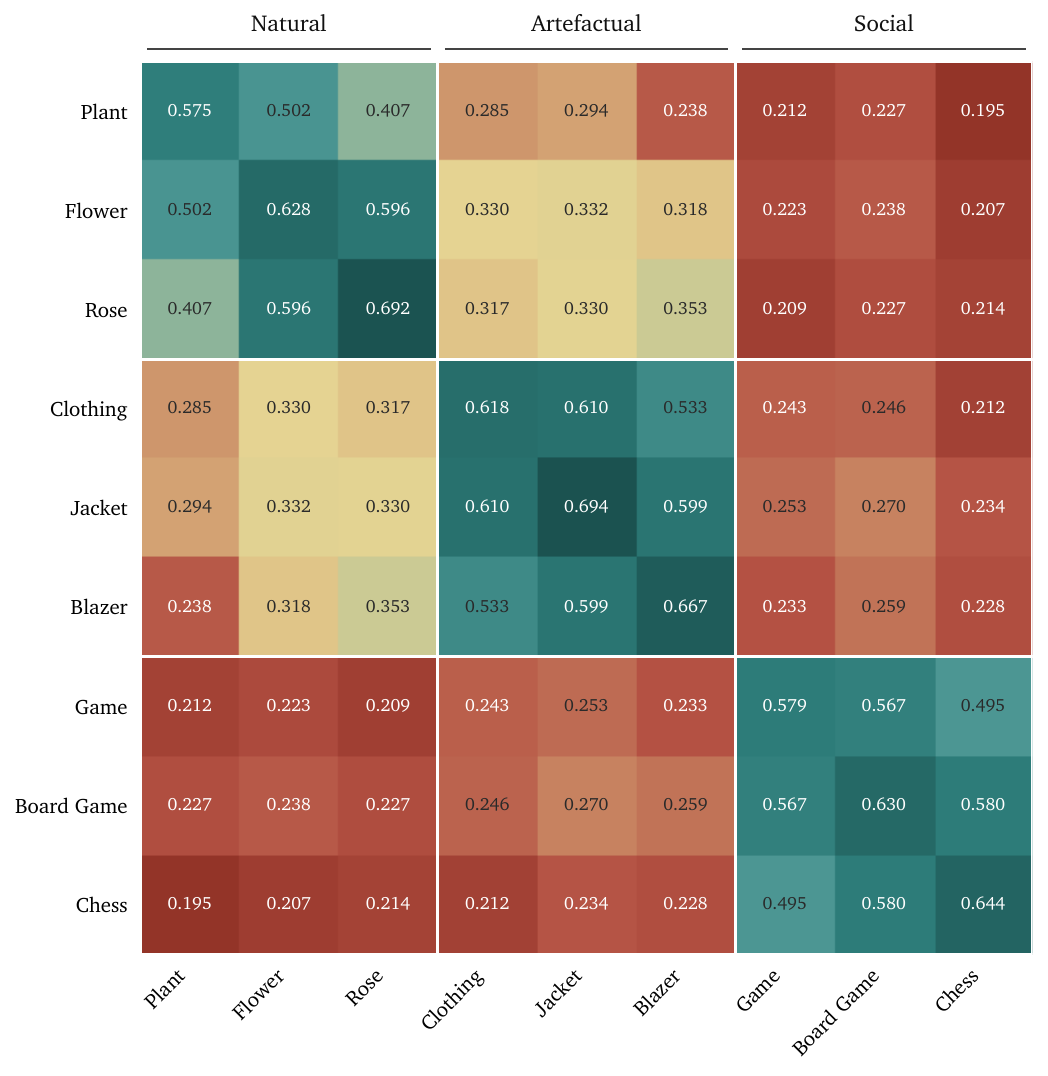}
\caption{Each cell contains the mean cosine similarity across all participant embedding pairs combining concept $b$ with concept $b'$. Diagonal cells reflect mean within-concept similarity across participants; off-diagonal cells reflect mean cross-concept similarity.}
\label{fig:similarity_matrix}
\end{figure*}

\begin{table*}[t]
\centering
\small
\caption{Item-level predictive validity under the primary encoder. Predicted and chosen counts are given for each response option. $N$ is the number of participants with a valid definition and a valid forced-choice response for that item. Chance baseline $= 0.500$.}
\label{tab:item_accuracy}
\begin{tabular}{llr rr rr r}
\toprule
 & & & \multicolumn{2}{c}{\textbf{Option A}} & \multicolumn{2}{c}{\textbf{Option B}} & \\
\cmidrule(lr){4-5}\cmidrule(lr){6-7}
\textbf{Concept} & \textbf{Item} & \textbf{$N$} & \textbf{Predicted} & \textbf{Chosen} & \textbf{Predicted} & \textbf{Chosen} & \textbf{Accuracy} \\
\midrule
Plant      & Item 1 & 116 &  31 &  41 &  85 &  75 & 0.603 \\
           & Item 2 & 116 & 113 &  95 &   3 &  21 & 0.828 \\
Flower     & Item 1 & 118 &  79 &  63 &  39 &  55 & 0.508 \\
           & Item 2 & 118 &  25 &  44 &  93 &  74 & 0.636 \\
Rose       & Item 1 & 117 &  26 &  62 &  91 &  55 & 0.538 \\
           & Item 2 & 117 &  91 & 100 &  26 &  17 & 0.701 \\
\addlinespace
Clothing   & Item 1 & 115 &  90 &  63 &  25 &  52 & 0.643 \\
           & Item 2 & 115 &  72 & 114 &  43 &   1 & 0.635 \\
Jacket     & Item 1 & 118 & 112 &  58 &   6 &  60 & 0.492 \\
           & Item 2 & 118 & 103 &  53 &  15 &  65 & 0.475 \\
Blazer     & Item 1 & 116 &  90 &  92 &  26 &  24 & 0.724 \\
           & Item 2 & 116 &  76 &  28 &  40 &  88 & 0.448 \\
\addlinespace
Game       & Item 1 & 117 & 115 &  71 &   2 &  46 & 0.590 \\
           & Item 2 & 117 &  73 &  98 &  44 &  19 & 0.632 \\
Board Game & Item 1 & 116 &  75 &  92 &  41 &  24 & 0.560 \\
           & Item 2 & 116 &  14 &  14 & 102 & 102 & 0.793 \\
Chess      & Item 1 & 118 & 115 & 112 &   3 &   6 & 0.924 \\
           & Item 2 & 118 &  86 &  86 &  32 &  32 & 0.610 \\
\midrule
\textit{Overall} & & \textit{2102} & & & & & \textit{0.630} \\
\bottomrule
\end{tabular}
\end{table*}

\subsection{Validity Tests}

\subsubsection{Structural Validity.} Structural validity was assessed via a 9$\times$9 concept-level mean cosine similarity matrix, where each cell $\bar{s}_{b,b'}$ contains the mean cosine similarity across all participant embedding pairs combining concept $b$ with concept $b'$. Diagonal cells ($b = b'$) reflect mean within-concept similarity, and off-diagonal cells ($b \neq b'$) reflect mean cross-concept similarity. 

This is a basic test of whether definition embeddings, with the shared sentence stem removed, capture their intended concepts and domains. Three properties are evaluated: (1) diagonal cells should exceed between-domain off-diagonal cells; (2) diagonal cells should exceed within-domain off-diagonal cells, a more demanding criterion given the semantic proximity of concepts within the same domain; and (3) within-domain off-diagonal similarity should exceed between-domain off-diagonal similarity.

\subsubsection{Predictive Validity.} A key function of concepts is their use as categories, where multiple entities can fall under a single category, some more typically than others \parencite{barsalou_instability_1987}. A predictive validity test was designed to evaluate whether the embedding space recovers typicality judgments in the forced-choice typicality task, an extensional use of concepts, from a participant's definition of that concept.

The two researcher-constructed options per item were encoded with the same encoder as participants' definitions, placing all three in a shared embedding space. The test assumes that if a participant's definition embedding accurately captures their concept representation, the description they find more typical should lie closer in embedding space than the alternative. For participant $i$, concept $b$, and item $k$, the predicted choice $\hat{y}_{ibk}$ is accordingly the description whose embedding has higher cosine similarity to the participant's definition embedding.

Accuracy is computed as the proportion of items for which $\hat{y}_{ibk}$ matches the observed choice $y_{ibk}$, pooled across participants and items. Confidence intervals (CIs; 95\%) are estimated via clustered bootstrap resampling of participants. Accuracy is assessed against a chance baseline of $p = 0.500$ and an upper bound defined by response consistency, the mean proportion of consistent responses across three repeated items.

\subsection{Robustness Tests}
Within the SBERT framework, numerous encoders are available, differing in architecture, pretraining corpora, fine-tuning objectives, and capacity, each inducing a distinct geometric space. Encoder choice therefore constitutes a non-trivial measurement decision. To assess whether the main results depend on it, the full pipeline was rerun under two alternative encoders specified in the pre-registration, \texttt{all-MiniLM-L12-v2} \parencite{wang_minilm_2020}, a lower-capacity distilled encoder, and \texttt{all-\allowbreak roberta-\allowbreak large-\allowbreak v1} \parencite{liu_roberta_2019}, a higher-capacity encoder, which together with the primary encoder span a meaningful range of capacities. 

Robustness was assessed at two stages of the pipeline. At the dyadic stage, prior to modeling, Spearman's $\rho$ was computed per concept between the pairwise dissimilarities $\delta_{ijb}$ of the primary encoder and those of each alternative. At the person stage, the CRE model was refit under each encoder, and Spearman's $\rho$ was computed per concept between the estimated random effects $\hat{a}_{ib}$ of the primary and each alternative encoder. Correlations are reported descriptively across concepts.

\section{Results}
\label{sec:results}
Results are reported in order of (i) validation, (ii) main analysis, and (iii) robustness checks.

\subsection{Validation}

\subsubsection{Structural Validity.} Results are reported in Figure~\ref{fig:similarity_matrix}. All three pre-registered criteria were satisfied. First, diagonal entries exceeded all between-domain entries, indicating higher within-concept than cross-domain similarity (e.g.,\allowbreak\ $\bar{s}_{\textit{rose}, \textit{rose}} \allowbreak= 0.692 > \allowbreak\bar{s}_{\textit{rose}, \textit{jacket}} \allowbreak= 0.330$). Second, diagonal entries exceeded within-domain off-diagonal entries (e.g.,\allowbreak\ $\bar{s}_{\textit{rose}, \textit{rose}} \allowbreak= 0.692 > \allowbreak\bar{s}_{\textit{rose}, \textit{flower}} \allowbreak= 0.596$), despite semantic proximity within domains. Third, within-domain similarities exceeded between-domain similarities across all three domains, recovering the expected domain partition.

\subsubsection{Predictive Validity.} Predictive validity was assessed by testing whether the embedding space predicts, via cosine similarity, which of two descriptions participants select in the forced-choice typicality task. Without any fitting or free parameters, the classifier achieved an overall accuracy of 0.630, 95\% CI [0.608, 0.651], significantly above chance and below the participant consistency ceiling of 0.893 (Table~\ref{tab:item_accuracy}; per-encoder accuracies in Table~\ref{tab:encoder_accuracy}). These results meet the pre-registered criterion, indicating that the embedding space systematically predicts within-participant extensional typicality judgments. 

In an exploratory, non-pre-registered test, within-participant accuracy was further compared against cross-participant accuracy, defined as the proportion of item-level comparisons, pooled across all participant pairs $(i, j)$, $i \neq j$, for which $j$'s predicted choice matched $i$'s observed response. Within-participant accuracy (0.630) exceeded cross-participant accuracy (0.606) by 0.024, 95\% CI [0.008, 0.040], indicating modest but reliable person-specific information not captured by a randomly selected other participant.

\subsection{Main Analysis}

\begin{figure*}[t]
\centering
\includegraphics[width=0.95\textwidth]{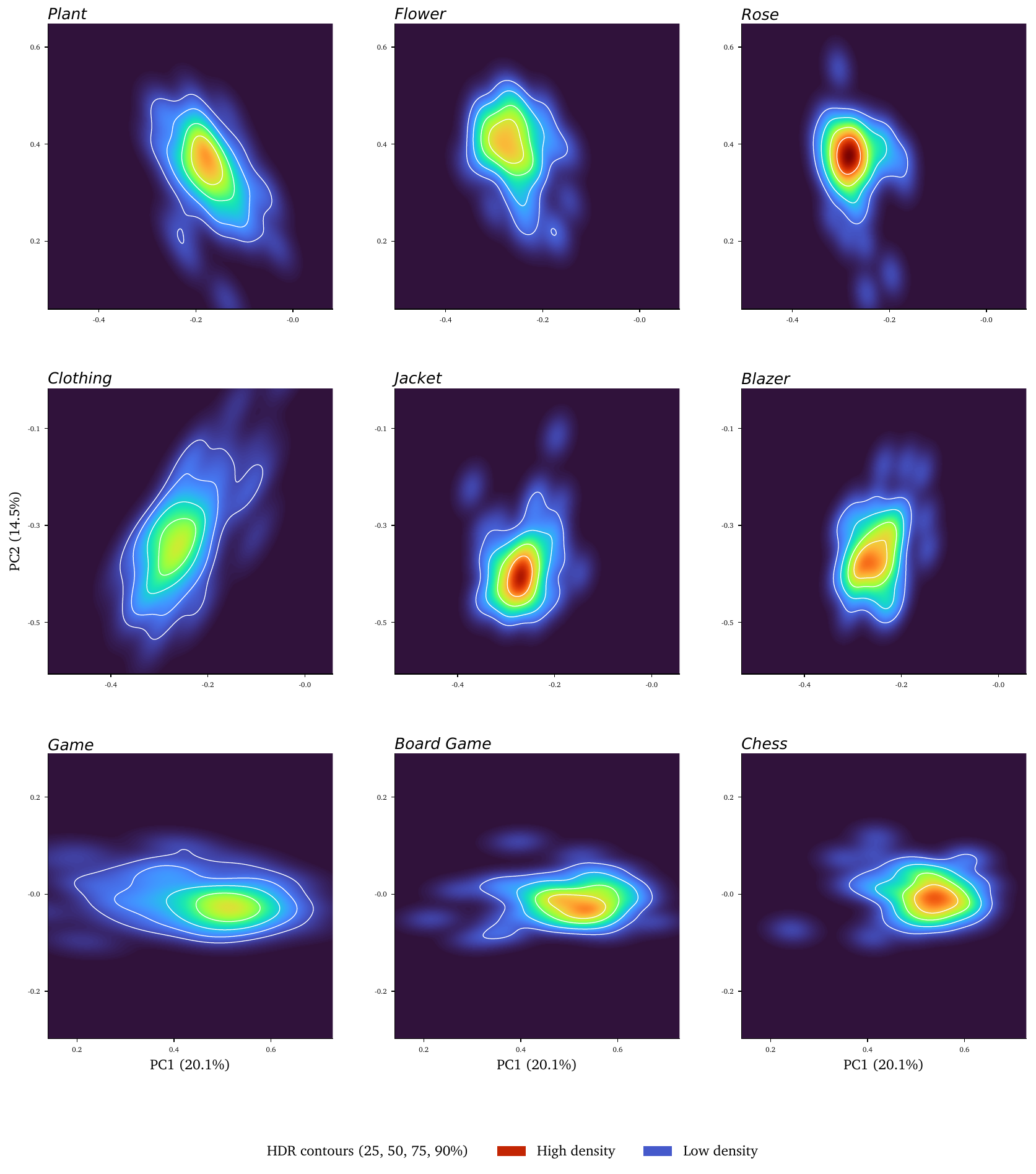}
\caption{Axis limits share a common half-width determined by the maximum data extent across triplets, enabling comparison of cloud size and concentration across panels. Each row is centered on its respective triplet; absolute positions across rows are therefore not comparable. Color intensity represents kernel density normalized to the global maximum, with warmer colors indicating higher density. HDR contours delineate the 25, 50, 75, and 90\% highest-density regions for each concept.
\label{fig:embedding_space}}
\end{figure*}

\begin{figure*}[t]
\centering
\includegraphics[width=0.95\textwidth]{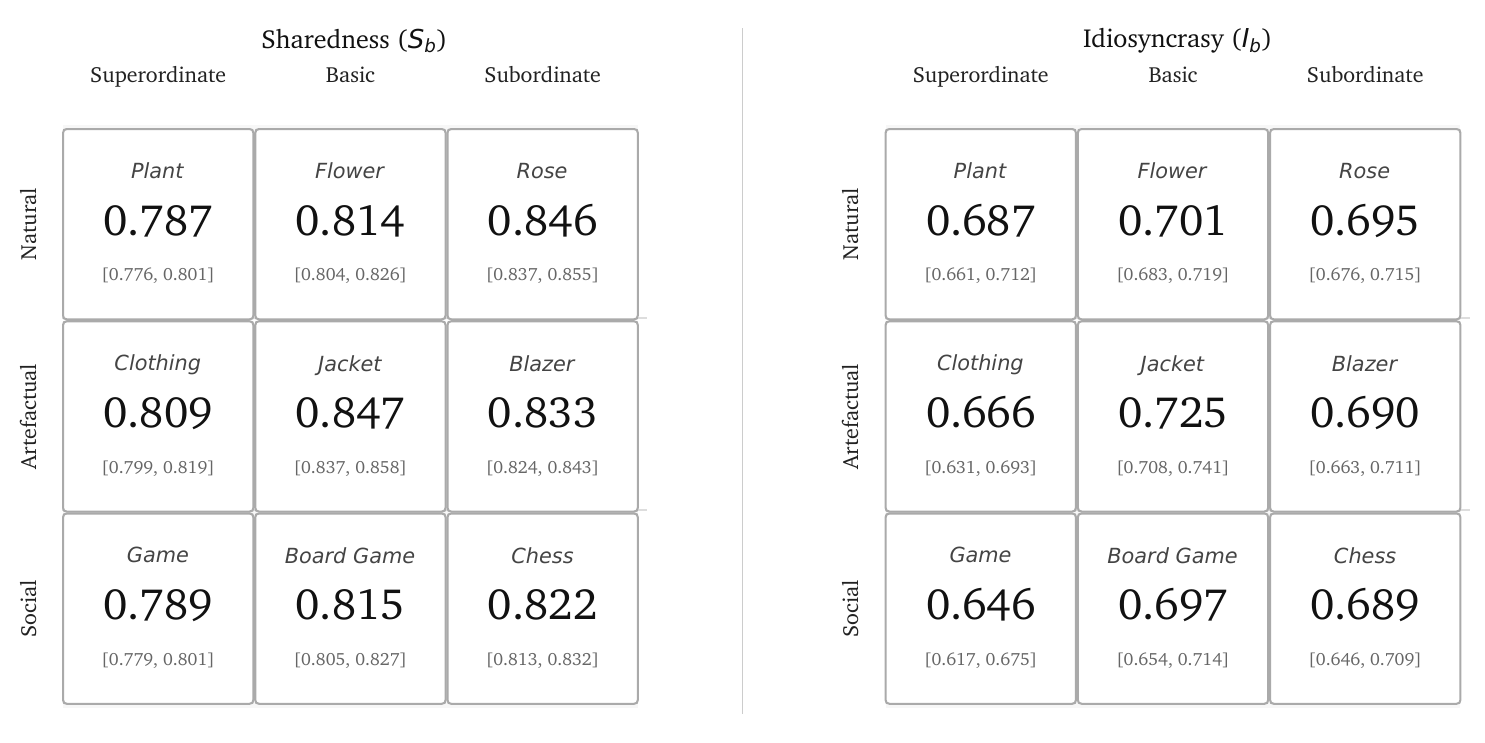}
\caption{Each cell displays the point estimate and 95\% bias-corrected bootstrap CI for $S_b$ and $I_b$ for a given concept.
\label{fig:main_analysis}}
\end{figure*}

Encoding all definitions within a shared embedding space allows variation to be examined visually, with spread comparable across participants and concepts. Figure~\ref{fig:embedding_space} projects the 768-dimensional distributions onto their first two principal components as kernel density estimates, one panel per concept, where tighter, more concentrated clouds indicate greater convergence across participants and broader, more diffuse clouds greater divergence. This becomes particularly apparent in the bottom row, where distributions narrow and become more centrally concentrated along the abstraction gradient. The projection accounts for 34.6\% of total variance and is provided for illustration only. All analyses operate on the full 768-dimensional vectors.

Concept-level estimates of $S_b$ and $I_b$ are reported in Figure~\ref{fig:main_analysis}. $S_b$ ranged from $0.787$ (\textit{plant}) to $0.847$ (\textit{jacket}). Prior work predicts that $S_b$ peaks at the basic level, with representations converging less strongly at subordinate \parencite{bellezza_reliability_1984-1} and superordinate categories \parencite{zhuang_characterization_2022}. Superordinate concepts showed the lowest $S_b$ within each triplet, corroborating this prediction. The taxonomic gradient, however, extended beyond the basic level in two of three domains, with $S_b$ increasing monotonically from superordinate to subordinate. The artefactual triplet was the exception, with \textit{jacket} $(0.847)$ exceeding \textit{blazer} $(0.833)$. No natural kind advantage in $S_b$ was observed.

$I_b$ estimates ranged from $0.646$ (\textit{game}) to $0.725$ (\textit{jacket}), indicating that individual distinctiveness accounts for 64.6\% to 72.5\% of total dyadic variance. CIs excluded zero across all concepts. A tentative pattern of higher $I_b$ at the basic level and lower $I_b$ in the social domain is observed, though CIs are largely overlapping.

The two estimands were empirically independent. Concepts with higher $S_b$ did not systematically show lower $I_b$, indicating that they capture distinct aspects of representational structure. 

\subsection{Robustness}
The full pipeline was replicated with two alternative encoders: \texttt{all-\allowbreak roberta-\allowbreak large-\allowbreak v1} and \texttt{all-\allowbreak MiniLM-\allowbreak L12-\allowbreak v2}. Dyadic-level correlations were high, ranging from $\rho = 0.742$ to $\rho = 0.837$ against the RoBERTa encoder and from $\rho = 0.742$ to $\rho = 0.835$ against the MiniLM encoder. Person-level correlations were comparable in magnitude, ranging from $\rho = 0.738$ to $\rho = 0.840$ and $\rho = 0.750$ to $\rho = 0.851$ respectively, indicating that the relative structure of pairwise dissimilarities and the relative positioning of individuals are broadly preserved across encoders.

Exploratorily, concept-level $S_b$ rankings were examined via Kendall's $\tau$, given that the main analysis recovered systematic differences across concepts. Rankings replicated strongly against both encoders ($\tau = 0.833$ and $\tau = 0.722$), indicating that the ordering of concepts by $S_b$ is robust to encoder choice.

\section{Discussion}
\label{sec:discussion}
The study introduces a framework for measuring variation in concept representations, validated against two criteria. Structural validity was satisfied throughout, showing that the embedding space recovers concept- and domain-level structure, even with the target word excluded prior to encoding. 

Cosine similarity in embedding space predicted forced-choice typicality judgments above chance but well below the participant-consistency ceiling. Two considerations suggest the result is more informative than it implies. First, the empirical link between intensions and extensions has proven notoriously difficult to establish \parencite{hampton_when_2016}, with \textcite{hampton_investigating_2020} describing it as ``a disturbing null result'' (p.~68). Second, the consistency ceiling likely overstates an attainable benchmark, as repeated-item agreement within a single session is susceptible to response recall. 

A further concern is item-level class imbalance. For several items, both classifier and participants overwhelmingly favored the same option, potentially inflating accuracy without implying a person-level intension--extension link. The exploratory cross-participant comparison partly addresses this: own-definition accuracy exceeded cross-participant accuracy, indicating a person-specific signal beyond item-level baselines, though the effect is modest.

$S_b$ estimates revealed non-overlapping CIs between several concepts, demonstrating the framework's sensitivity to reliably distinguish concepts on the mean scale. Across taxonomic level, $S_b$ increased monotonically in five of six within-domain comparisons, partly going against previous research predicting a basic-level peak \parencite{zhuang_characterization_2022}. Contrary to the prediction that perceptual grounding stabilizes natural kind representations \parencite{hacking_tradition_1991}, no natural kind advantage emerged. These findings should be interpreted as illustrative rather than confirmatory. They are based on a small set of concepts, and the convenience sample is student-heavy and geographically concentrated, limiting generalizability and potentially obscuring cross-country variation. Future work may adjudicate these patterns using broader concept sets and more principled sampling.

$I_b$ was high across concepts, with all CIs excluding zero. This provides the first continuous quantification of the share of dyadic variance attributable to individual distinctiveness, lending empirical grounding to the claim that individuals hold stable personal representations \parencite{hampton_investigating_2020}. Whether $I_b$ varies systematically across concepts remains open, as CIs largely overlapped, and calls for further research. $S_b$ and $I_b$ were empirically independent, capturing non-redundant aspects of representational structure. 

Results proved largely robust across encoders despite substantial differences in model size, among other properties, suggesting that the recovered variation is not an artefact of encoder-specific properties. The moderate computational requirements of the tested models further render the framework accessible to researchers without requiring extensive computational resources. Where greater resources permit, large language models adapted as text encoders represent a natural avenue for extension, having recently outperformed many SBERT-family models on benchmark tasks \parencite{wang_improving_2024}. 

\subsection{Limitations}
\label{sec:limitations}

Personal definitions were chosen for this demonstration because they elicit full compositional utterances. However, definitions may fail to capture the targeted mental structures, as they have been criticized for being overly generic and omitting obvious but non-distinguishing features \parencite{hampton_concepts_2006}. Despite instructions to describe their personal understanding, participants may also reproduce dictionary-like definitions, thereby tapping into a broadly shared, culturally consolidated layer of meaning. The framework is compatible with alternative elicitation formats, such as free association, feature listing, or episodic prompts, which may provide access to different aspects of concept representations.

Furthermore, a deeper epistemological issue in elicitation warrants acknowledgment. Eliciting concept representations by asking about their general meaning presupposes that concepts have stable content that can be studied outside the situations in which they are used \parencite{fodor_connectionism_1988}. This assumption has been challenged, most notably by Wittgenstein's later view that meaning lies in use \parencite{wittgenstein_philosophical_1958}, which could motivate future research to incorporate and systematically vary contextual information in elicitation.

For the analysis of the resulting textual traces, sentence embeddings are positioned as a principled measurement instrument, but they warrant several considerations. First, although robustness checks showed that variation is not an artefact of encoder choice, selecting among thousands of available encoders constitutes a non-trivial measurement decision. STS benchmarks provide a starting point but validate general similarity judgments rather than fine-grained discrimination among textual traces of the same concept \parencite{enevoldsen_mmteb_2025}.

Second, cosine-based comparisons of sentence embeddings may carry further limitations. Embedding spaces may exhibit non-uniform semantic resolution \parencite{valentini_investigating_2023, wang_unsupervised_2023}. Domains more extensively represented in training corpora may be encoded at finer granularity, such that equal cosine distances across concepts do not necessarily correspond to equal degrees of representational difference. Moreover, contextualized transformer representations often exhibit anisotropy, occupying a narrow cone in vector space that compresses the range of cosine similarity scores and inflates similarity among textual inputs \parencite{ethayarajh_how_2019}. One avenue for further development could be to move beyond cosine similarity between pooled sentence embeddings by decomposing textual traces at the token level and comparing them via a ColBERT-style late-interaction design \parencite{khattab_colbert_2020}, potentially yielding more fine-grained distance measures.

Third, it remains unclear what drives differences in embeddings. Standardized prompts, spelling-error correction, exclusion of nontopical responses, and high English proficiency requirements mitigate potential surface-level confounds but do not rule them out, representing a general limitation of embedding-based approaches.

Finally, encoders are trained predominantly on English and Western web text \parencite{bender_dangers_2021}, situating participant responses within a Western reference frame. Because embeddings inherit the biases of their training corpora \parencite{caliskan_semantics_2017}, this raises ethical considerations and warrants particular attention in cross-cultural applications and when studying normatively sensitive concepts.

\section{Conclusion}
\label{sec:conclusion}
The measurement framework demonstrates broader methodological promise. It operates on short textual responses and yields stable individual-level estimates in samples typical of survey research. Unlike closed-ended survey formats, modeling standardized elicited textual responses may offer a novel, so far underused empirical window into mental structures, proceeding bottom-up without requiring researchers to define relevant dimensions a priori. Several methodological avenues for further development were outlined, including alternative elicitation formats, LLM-based encoders, and alternatives to cosine similarity and single-vector pooling, such as ColBERT-style late interaction.

For researchers interested in applying the framework, this opens a range of questions that can be addressed by identifying directed hypotheses and testing them across broader concept sets. The variation captured by the framework can also be further investigated statistically. One pertinent question, for instance, could be to move beyond uncovering interindividual differences and investigate which social variables explain sharedness in concept representations.

\section*{Code and Data Availability}
A fully anonymized dataset retaining minimal and categorized sociodemographic  information and survey responses, along with a reproducible analysis codebook, are available at \osfdata.

\section*{Author Contributions}
S.J.D. led the project and was responsible for conceptualization, methodology, investigation (data collection), software, formal analysis, visualization, and writing (original draft).
D.B. provided supporting input on the philosophical framing of the argument (conceptualization, supporting).

\section*{Acknowledgments}
This work was conducted in part during a research stay of S.J.D.\ at the Institut Jean Nicod, École normale supérieure -- PSL (February--May 2026).

\printbibliography

\onecolumn
\appendix
\section{Supplementary Tables}
\label{sec:appendix}
\appendix
\setcounter{table}{0}
\renewcommand{\thetable}{A\arabic{table}}

This appendix reports the sample composition (Table~\ref{tab:sociodemographic}), the forced-choice items (Table~\ref{tab:closedend}), item-level accuracies under all three encoders (Table~\ref{tab:encoder_accuracy}), and the definitions excluded prior to analysis (Table~\ref{tab:excluded_definitions}).

\begin{table}[!h]
\small
\begin{minipage}[t]{0.52\textwidth}
\vspace{0pt}\centering
\setlength{\tabcolsep}{5pt}
\caption{Sociodemographic characteristics of the sample ($N = 125$). Values are frequencies with percentages in parentheses unless otherwise indicated. Age is reported as mean with standard deviation; range = 18--77. Percentages may not sum to 100 due to rounding.}
\label{tab:sociodemographic}
\begin{tabular}{@{}>{\raggedright\arraybackslash}p{2.1cm}lr@{}}
\toprule
\textbf{Variable} & \textbf{Category} & \textbf{$n$ (\%) / $M$ (\textit{SD})} \\
\midrule
Age &  & 25.9 (9.1) \\
\addlinespace
Gender & Woman & 73 (58.4) \\
 & Man & 50 (40.0) \\
 & Prefer to self-describe & 1 (0.8) \\
 & Prefer not to say & 1 (0.8) \\
\addlinespace
Native English speaker & Yes & 24 (19.2) \\
 & No & 101 (80.8) \\
\addlinespace
Education & No formal qualification & 4 (3.2) \\
 & Compulsory schooling & 4 (3.2) \\
 & Upper secondary & 20 (16.0) \\
 & Vocational & 8 (6.4) \\
 & Bachelor's degree & 57 (45.6) \\
 & Master's degree & 27 (21.6) \\
 & Doctorate & 4 (3.2) \\
 & Other & 1 (0.8) \\
\addlinespace
Employment & Employed full-time & 28 (22.4) \\
 & Employed part-time & 9 (7.2) \\
 & Self-employed & 4 (3.2) \\
 & Student & 75 (60.0) \\
 & Unemployed & 2 (1.6) \\
 & Not in labour force & 1 (0.8) \\
 & Retired & 1 (0.8) \\
 & Other & 5 (4.0) \\
\addlinespace
Country of residence & Germany & 38 (30.4) \\
 & France & 20 (16.0) \\
 & Italy & 16 (12.8) \\
 & Denmark & 16 (12.8) \\
 & South Africa & 9 (7.2) \\
 & Spain & 5 (4.0) \\
 & Belgium & 3 (2.4) \\
 & United Kingdom & 3 (2.4) \\
 & United States & 3 (2.4) \\
 & Netherlands & 2 (1.6) \\
 & Other / unspecified & 9 (7.2) \\
\bottomrule
\end{tabular}
\end{minipage}\hfill
\begin{minipage}[t]{0.44\textwidth}
\vspace{0pt}\centering
\captionsetup{justification=raggedright}
\setlength{\tabcolsep}{5pt}
\caption{Item-level accuracy under the primary encoder (mpnet: \texttt{all-mpnet-base-v2}) and the two pre-registered robustness encoders (RoBERTa: \texttt{all-roberta-large-v1}; MiniLM: \texttt{all-MiniLM-L12-v2}). Accuracy is the proportion of forced-choice responses for which the predicted option matched the observed choice, pooled across participants. Item-level discrepancies between encoders were largest for Clothing~2, where RoBERTa and MiniLM substantially exceeded mpnet accuracy, and for Blazer~1 and Board Game~1, where the pattern reversed. Chance baseline $= 0.500$.}
\label{tab:encoder_accuracy}
\begin{tabular}{@{}lrrr@{}}
\toprule
\textbf{Item} & \textbf{mpnet} & \textbf{RoBERTa} & \textbf{MiniLM} \\
\midrule
Plant 1       & 0.603 & 0.560 & 0.603 \\
Plant 2       & 0.828 & 0.828 & 0.741 \\
Flower 1      & 0.508 & 0.525 & 0.559 \\
Flower 2      & 0.636 & 0.627 & 0.610 \\
Rose 1        & 0.538 & 0.521 & 0.624 \\
Rose 2        & 0.701 & 0.709 & 0.726 \\
\addlinespace
Clothing 1    & 0.643 & 0.652 & 0.661 \\
Clothing 2    & 0.635 & 0.948 & 0.939 \\
Jacket 1      & 0.492 & 0.500 & 0.534 \\
Jacket 2      & 0.475 & 0.576 & 0.585 \\
Blazer 1      & 0.724 & 0.767 & 0.276 \\
Blazer 2      & 0.448 & 0.707 & 0.603 \\
\addlinespace
Game 1        & 0.590 & 0.615 & 0.590 \\
Game 2        & 0.632 & 0.376 & 0.444 \\
Board Game 1  & 0.560 & 0.328 & 0.466 \\
Board Game 2  & 0.793 & 0.672 & 0.828 \\
Chess 1       & 0.924 & 0.847 & 0.585 \\
Chess 2       & 0.610 & 0.644 & 0.636 \\
\midrule
\textit{Overall} & \textit{0.630} & \textit{0.633} & \textit{0.611} \\
\bottomrule
\end{tabular}
\end{minipage}
\end{table}

\begin{table}[p]
\centering
\small
\renewcommand{\arraystretch}{1.08}
\caption{Forced-choice typicality items. The task consisted of 18 items, two per concept. For each item, participants selected the description that most closely aligned with their own understanding of the concept. Each pair of descriptions contrasts two options differing along dimensions relevant to that concept.}
\label{tab:closedend}
\begin{tabularx}{\textwidth}{@{}l l c X@{}}
\toprule
\textbf{Domain} & \textbf{Concept} & \textbf{Item} & \textbf{Descriptions} \\
\midrule
Natural & Plant & 1 & (A) A tall oak tree rising from the ground in a wild forest. \\
 & & & (B) A small ornamental orchid kept indoors in a pot. \\
\addlinespace[2pt]
 & & 2 & (A) A dense green bush with broad leaves. \\
 & & & (B) A blooming bush covered in red, orange, and pink dahlias. \\
\addlinespace[4pt]
 & Flower & 1 & (A) A carefully cultivated tulip growing in a garden. \\
 & & & (B) A wild dandelion growing in a field. \\
\addlinespace[2pt]
 & & 2 & (A) A white lily placed on a grave at a funeral. \\
 & & & (B) A bouquet of sunflowers given to celebrate a birthday. \\
\addlinespace[4pt]
 & Rose & 1 & (A) A flowering bush with sharp thorns growing in a garden. \\
 & & & (B) A traditional symbol of love given on Valentine's Day. \\
\addlinespace[2pt]
 & & 2 & (A) A bloom with dark red petals. \\
 & & & (B) A bloom with bright white petals. \\
\midrule
Artefactual & Clothing & 1 & (A) An item used to protect the body from cold or weather. \\
 & & & (B) An item chosen to signal personal style or identity. \\
\addlinespace[2pt]
 & & 2 & (A) A T-shirt worn on the upper body. \\
 & & & (B) A pair of shoes worn on the feet. \\
\addlinespace[4pt]
 & Jacket & 1 & (A) Thick, padded outerwear that keeps you warm in snow and frost. \\
 & & & (B) Lightweight outerwear thrown over a t-shirt on a breezy day. \\
\addlinespace[2pt]
 & & 2 & (A) A Gore-Tex hiking jacket worn through wind and rain on a mountain trail. \\
 & & & (B) A sleek leather jacket chosen to reflect a personal aesthetic. \\
\addlinespace[4pt]
 & Blazer & 1 & (A) Something worn to signal authority and competence in a high-stakes corporate board meeting. \\
 & & & (B) Something worn to a lighthearted garden party on a summer afternoon. \\
\addlinespace[2pt]
 & & 2 & (A) An oversized, retro piece with a check pattern for everyday casual wear. \\
 & & & (B) A tailored, navy-blue attire with a fitted cut for formal occasions. \\
\midrule
Social & Game & 1 & (A) A sport like volleyball played outdoors with multiple players. \\
 & & & (B) A logic puzzle like Sudoku solved alone in silence. \\
\addlinespace[2pt]
 & & 2 & (A) A game of Uno played with a physical deck of cards around a table. \\
 & & & (B) A game of Tetris played on a screen in a virtual digital environment. \\
\addlinespace[4pt]
 & Board Game & 1 & (A) Something competitive where players eliminate their opponents, with only one person winning. \\
 & & & (B) Something cooperative where players coordinate their actions toward a shared goal to win as a team. \\
\addlinespace[2pt]
 & & 2 & (A) A round of Twister where players stretch their bodies across a spotted mat without falling. \\
 & & & (B) A round of Connect Four where players think through patterns and plan their moves carefully. \\
\addlinespace[4pt]
 & Chess & 1 & (A) A strategic game won by analyzing positions carefully and calculating sequences of moves several steps ahead. \\
 & & & (B) A creative game won by intuitively spotting original ideas and inventing unexpected combinations. \\
\addlinespace[2pt]
 & & 2 & (A) A game played in formal competitions that emphasize concentration and serious intellectual skill. \\
 & & & (B) A game played informally in parks or caf\'{e}s as a relaxed and social pastime. \\
\bottomrule
\end{tabularx}
\end{table}

\begin{table}[p]
\centering
\small
\renewcommand{\arraystretch}{1.15}
\caption{Definitions excluded prior to analysis, reproduced verbatim. Spelling errors present in the original responses are retained.}
\label{tab:excluded_definitions}
\begin{tabularx}{\textwidth}{@{}l >{\raggedright\arraybackslash}X >{\raggedright\arraybackslash}p{0.3\textwidth}@{}}
\toprule
\textbf{Concept} & \textbf{Definition (verbatim)} & \textbf{Exclusion criterion} \\
\midrule
Blazer     & \textit{A blazer is I was really not sure what it was when I was answering cause my english was not englishing please dont take this concept into account :)} & Explicit request to not be included \\
\addlinespace
Plant      & \textit{A plant is boring. If it's very green it can be pretty but I can't water them. I always forget. And that's it.} & Largely affective associations \\
\addlinespace
Rose       & \textit{A rose is pink flower, with soft petals. ``C'est une folie de haïr toutes les roses parce qu'une épine vous a piqué''} & Non-English content \\
\addlinespace
Clothing   & \textit{Clothing is something you wear.} & Minimal definitional content; character minimum circumvented because of drop-out \\
\addlinespace
Board Game & \textit{A board game is an event where you can really easily be mad at your friends.} & Minimal definitional content \\
\addlinespace
Chess      & \textit{Chess is a very funny game. Especially in competitions. I love the book ``the chess player'' by Zweig which tells the story of a schinzophren man.} & No definitional content about the target concept \\
\addlinespace
Game       & \textit{A game is a game is sometimes an event where you play or compete with friends, opponents event competer ot sometime you can play buy your self} & Response disfluent to the point that definitional intent cannot be reliably extracted \\
\addlinespace
Rose       & \textit{A rose is beautiful smell and disgner we had in our nature most represent our love we share in the world those colors bring back hope} & Response disfluent to the point that definitional intent cannot be reliably extracted \\
\addlinespace
Clothing   & \textit{Clothing is wear we use to cover our body in our daily basis we can't leave without it we need clothes to hide it} & Response disfluent to the point that definitional intent cannot be reliably extracted \\
\addlinespace
Jacket     & \textit{A jacket is a cloth prevent us to get cold whatever we go we wear it if it's cold we go out for the help of it so we glad to have our jacket} & Response disfluent to the point that definitional intent cannot be reliably extracted \\
\addlinespace
Board Game & \textit{A board game is is games we play on play station those games like candy crush who were created online to relax our minds} & Response disfluent and unrelated to the target concept \\
\addlinespace
Plant      & \textit{A plant is something that you put in and expect a postive results out, it can be planting a positive mindset or planting veggies what matters is the feedback.} & No definitional content, largely metaphorical \\
\addlinespace
Chess      & \textit{Chess is a type of food that gives a unique taste or look to the food and it's very fast to cook when one is in a hury for something.} & Unrelated to the target concept \\
\bottomrule
\end{tabularx}
\end{table}

\end{document}